\documentclass[sigconf]{acmart}

\AtBeginDocument{%
  }

\copyrightyear{2025}
\acmYear{2025}
\setcopyright{cc}
\setcctype{by-nc-sa}
\acmConference[MMSports '25]{Proceedings of the 8th International ACM Workshop on Multimedia Content Analysis in Sports}{October 27--28, 2025}{Dublin, Ireland}
\acmBooktitle{Proceedings of the 8th International ACM Workshop on Multimedia Content Analysis in Sports (MMSports '25), October 27--28, 2025, Dublin, Ireland}
\acmDOI{10.1145/3728423.3759410}
\acmISBN{979-8-4007-1835-9/2025/10}

\usepackage{multirow} 
\usepackage{subcaption}
\usepackage{comment}
\usepackage{pifont}
\usepackage{graphicx}
\usepackage{amsfonts}
\usepackage{dsfont}
\usepackage{hyperref}
\usepackage{enumitem}
\graphicspath{ {./figures/} }

\begin{document}

\title{SoccerHigh: A Benchmark Dataset for Automatic Soccer Video Summarization}

\author{Artur Díaz-Juan}
\affiliation{
  \institution{Universitat Pompeu Fabra}
  \city{Barcelona}
  \country{Spain}
}
\email{artur.diaz@upf.edu}

\author{Coloma Ballester}
\affiliation{
  \institution{Universitat Pompeu Fabra}
  \city{Barcelona}
  \country{Spain}
}
\email{coloma.ballester@upf.edu}

\author{Gloria Haro}
\affiliation{
  \institution{Universitat Pompeu Fabra}
  \city{Barcelona}
  \country{Spain}
}
\email{gloria.haro@upf.edu}

\renewcommand{\shortauthors}{Artur Díaz-Juan, Coloma Ballester \& Gloria Haro}

\begin{abstract}
Video summarization aims to extract key shots from longer videos to produce concise and informative summaries. One of its most common applications is in sports, where highlight reels capture the most important moments of a game, along with notable reactions and specific contextual events. Automatic summary generation can support video editors in the sports media industry by reducing the time and effort required to identify key segments. However, the lack of publicly available datasets poses a challenge in developing robust models for sports highlight generation. In this paper, we address this gap by introducing a curated dataset for soccer video summarization, designed to serve as a benchmark for the task. The dataset includes shot boundaries for 237 matches from the Spanish, French, and Italian leagues, using broadcast footage sourced from the SoccerNet dataset. Alongside the dataset, we propose a baseline model specifically designed for this task, which achieves an F1 score of $0.3956$ in the test set. Furthermore, we propose a new metric constrained by the length of each target summary, enabling a more objective evaluation of the generated content.
The dataset and code are available at \url{https://ipcv.github.io/SoccerHigh/}.
\end{abstract}




\maketitle

%% Introductiom
\section{Introduction}
The generation of highlights from sports events is a common practice in the media industry. In today’s era of social networks, every organization—from individual players to entire leagues—maintains an online presence to maximize visibility. Audiences now prefer fast on-demand content, making access to key moments of a match almost as important as the complete match itself~\cite{nielsensports2022fans}.

Video summarization addresses the need to condense long footage into short videos that capture the most important events. In soccer, this involves selecting and compiling significant plays, such as goals and other pivotal moments, into a coherent summary. This process combines objective actions with subjective editorial choices, making it both technical and creative.

A key challenge in soccer video summarization is the limited availability of annotated public datasets that link complete matches to their corresponding highlight summaries. To address this gap, we introduce a curated dataset specifically designed for this task, derived from official league-produced summaries and aligned with the corresponding segments from full-match broadcast videos. To the best of our knowledge, this is the first public dataset for soccer video summarization.  The dataset comprises 237 paired full-match and summary videos covering three major European leagues: Spanish, Italian, and French. This diversity captures a range of editorial styles, emphasizing that video summarization extends beyond simply compiling in-game events.
Importantly, the dataset requires only the original broadcast video as input, facilitating the development of automatic summarization models and significantly reducing the manual effort involved in highlight production.

To minimize the need for manual dataset annotation, we propose a semi-automated approach to facilitate the identification of interesting scenes to be included in the summary. This method consists of two main stages: an initial shot segmentation of the summary video, followed by alignment with the corresponding segments in the full-match broadcast video. A final manual refinement step is performed to ensure consistent and accurate pair alignment.

In addition, we present a baseline model for the task of video summarization in soccer matches, intended as a foundation for the development of novel approaches in the field. This baseline provides reference metrics that enable the evaluation of models without constraining the duration of the produced summary.

\noindent
Accordingly, our main contributions are as follows:
\begin{itemize}[noitemsep, topsep=0pt]
    \item A curated dataset for soccer video summarization, aligned with full-match videos from SoccerNet \cite{deliege2021soccernet} at the shot level and reflecting diverse editorial styles, with standard training, validation, and test splits for comparison.
    \item A semi-automated annotation pipeline that generates accurate summary-to-broadcast alignments while significantly reducing annotation effort.
    \item A baseline model for soccer video summarization that operates directly on full-match broadcast videos and serves as a reference for future research.
    \item An objective evaluation metric constrained by the ground-truth summary length.
\end{itemize}

%% Related work
\section{Related work}\label{section:relatedwork}

\paragraph{\textbf{Video Summarization Datasets}}

TVSum \cite{song2015tvsum} and SumMe \cite{gygli2014creating} are widely used benchmarks and are considered standards for the video summarization task. TVSum contains 50 YouTube videos spanning 10 diverse categories (e.g., news, how-to, sports, documentaries), with 5 videos per category. Each video is segmented into shots and manually annotated with importance scores. SumMe consists of 25 raw user-generated videos covering a variety of everyday activities (e.g., holidays, sports, cooking), each annotated with multiple ground truth summaries by different human annotators. The videos range from approximately 1 to 6 minutes, with summary lengths typically between 5\% and 15\% of the original video duration.

Other datasets provide frame- or shot-level annotations for summarization, such as \cite{zhu2023topic, lei2107qvhighlights, zeng2016title, sun2014ranking}, while others incorporate textual transcripts for multimodal summarization tasks, including \cite{he2023align, miech2019howto100m, sanabria2018how2}.

Although existing datasets have enabled effective approaches to video summarization, they primarily cover domains such as general web content and instructional videos, which differ significantly from the task of summarizing soccer matches. Although some of them include sports-related content, they do not adequately capture the unique dynamics of a soccer game. In the case of soccer matches videos, we face specific challenges such as long video durations, instant replays, and high visual redundancy, as the soccer pitch remains the predominant background throughout the footage.

\paragraph{\textbf{Soccer Video Understanding Datasets}}

In soccer video analysis, the authors in \cite{giancola2018soccernet} introduced the SoccerNet dataset, comprising $500$ annotated soccer matches for action spotting across three categories: goals, cards, and substitutions. This dataset was then extended in \cite{deliege2021soccernet} to cover 17 distinct soccer actions, with an additional 50 games. The extended version also introduced new tasks, such as camera shot segmentation, boundary detection, and replay grounding, all of which are highly relevant to the development of effective soccer video summarization techniques. Furthermore, the third release, SoccerNet-v3 \cite{cioppa2022scaling}, includes spatial annotations and cross-view correspondences, providing rich localized information crucial for fine-grained video understanding.

Alternative datasets that are based on and complement SoccerNet include \cite{gao2020automatic}, SoccerDB \cite{jiang2020soccerdb}, and SoccerReplay-1988 \cite{rao2024towards}. In \cite{gao2020automatic}, the authors present a dataset comprising $460$ soccer matches focused on the extraction of video highlights. While it provides extensive event-recognition annotations enabling the detection of key moments such as goals and fouls, it lacks explicit annotations for summarization tasks. Consequently, experiments mainly address event recognition and localization rather than generating concise, human-like highlight summaries.
SoccerDB~\cite{jiang2020soccerdb} is a large-scale, publicly available multitask dataset that supports object detection, action recognition, temporal action localization, and highlight detection. It provides rich frame-level annotations and enables end-to-end video understanding across diverse soccer scenarios, based on video segments sampled from 346 high-quality soccer matches.
Similarly, SoccerReplay-1988~\cite{rao2024towards}, which is also publicly available, includes $1988$ match broadcasts from major European competitions between $2014$ and $2024$. Matches are annotated with second-level event labels across $24$ fine-grained classes (e.g., goals, penalties, VAR checks) and time-stamped natural language commentary generated via automated processing and LLM summarization. This multimodal data supports tasks like event classification, commentary generation, and multiview foul recognition.

More recently, SoccerSum \cite{sarkhoosh2024soccersum} introduced $750$ annotated frames from the Norwegian league ($2021$–$2023$), curated from $41$ video sequences. This dataset contains annotations for eight detection classes (player, goalkeeper, referee, ball, logo, penalty mark, corner flag post, and goal net) and two segmentation classes (penalty box and goal box). It is designed to support multiple downstream tasks, including automatic summarization, and is freely available for research purposes.
 
Although some of these datasets report results in tasks such as highlight detection or automatic summarization, they do not include dedicated annotations for these purposes. Instead, they rely on annotations for related tasks, such as action recognition, and derive summaries by detecting specific events. As a result, they often overlook the contextual elements that are essential for constructing a cohesive, viewer-friendly summary that reflects the narrative flow of a match.

\paragraph{\textbf{Video Summarization models}}

Most general video summarization models rely primarily on the previously mentioned datasets, TVSum and SumMe. Consequently, these approaches are typically based on supervised learning, taking advantage of available summary annotations to guide the detection of important segments. Models such as \cite{guo2025cfsum, xu2024mh, jiang2024mct, zhu2023topic, he2023align, chen2022video, li2021exploring, apostolidis2021combining, narasimhan2021clip, ji2019video} commonly incorporate prediction heads designed to identify key shots by means of saliency scores computed from the original videos.

Some of the most recent approaches \cite{guo2025cfsum, xu2024mh, jiang2024mct, he2023align, narasimhan2021clip} incorporate multiple modalities in addition to visual features, with the aim of achieving a more robust summarization by leveraging audio, speech, or text. Specifically, \cite{xu2024mh} and \cite{narasimhan2021clip} introduce queries into the video stream using the CLIP backbone \cite{radford2021learning}, while \cite{he2023align} incorporates the full audio transcript. Furthermore, \cite{guo2025cfsum} and \cite{jiang2024mct} take this step further by integrating both the audio signal and linking textual descriptions extracted from the videos.

Unsupervised approaches are less common than supervised ones, but some research has demonstrated effective summary generation without relying on annotated data. These models typically use TVSum and SumMe only during the evaluation stage. For example, contrastive learning is employed in \cite{sosnovik2023learning}, hierarchical LSTM networks with attention mechanisms are used in \cite{lin2022deep}, and a generative adversarial network is introduced in \cite{apostolidis2020ac} and used in \cite{gansum_gl_rpe} where they also sample input both globally and locally. Despite the lack of supervision, these methods achieve results that are competitive with those of their supervised counterparts. They generate summaries using greedy algorithms such as top-$k$ frame selection or the Knapsack algorithm~\cite{mathews1896partition}, which help in selecting the most representative segments under length constraints. 

Focusing on automatic soccer video summarization, earlier approaches have primarily tackled the task of highlight generation using architectures based on LSTMs \cite{scotti2019sferanet, sanabria2020profiling, sanabria2021hierarchical, sanabria2022multi, agyeman2019soccer}. These works laid the groundwork for soccer-specific summarization models with the aim of selecting the most important shots of a game. In \cite{scotti2019sferanet}, the authors use audio and transcriptions of $369$ matches in the Italian league to train a multimodal model without visual information, followed by manual editing to produce the final highlights. In \cite{sanabria2020profiling}, a rule-based system is proposed to identify interesting actions for customized summaries, using $70$ English and $20$ French league matches. The same authors further explore hierarchical multimodal architectures in \cite{sanabria2021hierarchical, sanabria2022multi}, leveraging attention mechanisms and event metadata to identify the most relevant actions, benchmarking against human-created summaries.

A more recent approach \cite{sarkhoosh2024multimodal} introduces the concept of social media-oriented summarization, generating short clips specifically designed for online sharing. This work is built on their own dataset, SoccerSum \cite{sarkhoosh2024soccersum}, and proposes a framework to automate goal-centric summaries by extracting information from multiple modalities and external sources.

As highlighted in the discussion of the dataset, most existing summarization models for soccer rely heavily on action recognition or event detection, rather than being designed explicitly for the video summarization task. This underscores the absence of a dedicated benchmark for comprehensive soccer match summarization, which our work aims to address.

%% Dataset
\section{Proposed Dataset} \label{section:Dataset}

Video summarization in the context of soccer involves generating a shorter video that captures key moments of a match, along with notable reactions and specific contextual events. These summaries typically combine game highlights, extracted from the most significant plays, with subjective shots (e.g., close-ups of coaches, crowd reactions) that enrich the overall viewing experience.

We describe in Sections~\ref{subsection:pipeline_dataset} and \ref{subsection:description_dataset} the creation of a curated dataset that pairs each full-match broadcast video with a professionally crafted summary produced by media and soccer experts. Then, Section~\ref{subsection:experiments_dataset} presents an experimental analysis allowing to further establish and fix the proposed summary annotation process.

The broadcast videos used in our dataset are sourced from the publicly available SoccerNet dataset \cite{deliege2021soccernet}, selecting matches that have the corresponding official summaries available online. Among the leagues covered by SoccerNet, only the official channels of the Spanish, French, and Italian leagues consistently provide such summaries for the available seasons. Consequently, our dataset comprises 237 games from these three leagues, each paired with a SoccerNet broadcast video and its corresponding official summary retrieved from the Internet.

\subsection{Pipeline} \label{subsection:pipeline_dataset}

To establish correspondences, we begin by detecting boundaries in the summary video to identify the individual shots it comprises. For this purpose, we apply a shot boundary detector (SBD) that produces candidate segment cuts. We explore two different strategies. The first is a custom from-scratch k-Nearest Neighbors (kNN) frame comparison, while the second leverages a pretrained SBD network, namely TransNetv2 \cite{soucek2020transnetv2}.

Our custom kNN-based method involves frame-by-frame feature comparison using descriptors extracted with DINO \cite{caron2021emerging}, trained on the Google Landmarks v2 dataset \cite{weyand2020google} for image retrieval tasks. Each frame is compared against all others to retrieve its \textit{k} nearest neighbors (with $\textit{k} = 10$). We then analyze the temporal indices from similar images proposed by the algorithm. After filtering isolated temporal outliers, if non-consecutive frames exist among the kNN set, we infer a boundary at the last index of the consecutive frame sequence. This approach ensures that all visually similar and temporally adjacent frames are grouped in the same shot. Since each scene in the summary may contain various types of frame, we intentionally oversegment the video to avoid missing any relevant shot during the alignment phase.

In contrast, TransNetv2 is applied directly to all summary frames, using two decision thresholds: $0.5$ and $0.05$. The threshold of $0.5$, as suggested by the authors, provides optimal boundary selection. Meanwhile, the $0.05$ threshold is used to induce oversegmentation, similar to the kNN strategy, ensuring that no shots are lost in the subsequent alignment stage.

Then, let the summary shots detected by the SBD algorithm be denoted by $s_i$, $i=1, ..., N_s$, where $N_s$ is the total number of shots. The visual content of each frame is represented by a feature vector $\mathbf{f}_k$, $k=1, ..., N_f$, where $N_{f}$ is the total number of frames extracted from the summary video. Each shot $s_i$ spans $l_i$ frames and contains the sequence of features corresponding to the $l_i$ frames within the shot:
\begin{equation}
s_i = \{\mathbf{f}_{k_i}, \mathbf{f}_{k_{i} + 1}, \dots, \mathbf{f}_{k_i + l_i - 1}\}, \text{with} 
\ k_{i} = 1 +
\begin{cases}
 \sum_{m=1}^{i-1} l_m, & \text{if } i>1 \\
0 & \text{otherwise.}
\end{cases}
\label{eq:summary_shots}
\end{equation}

Next, we perform an alignment stage using the feature representations extracted from each frame. Each summary shot $s_i$, as defined in Equation~\eqref{eq:summary_shots}, is compared against segments of the full broadcast video that are represented as a sequence of frame-level feature vectors $\mathbf{g}_j$, $j=1, ..., N_b$, where $N_b$ is the number of frames in the broadcast video. 

In this stage, we experiment with various backbone networks to extract frame-level feature representations, both for $\mathbf{f}_k$ and $\mathbf{g}_j$. Specifically, we evaluate the following models: ResNet-152~\cite{he2016deep} pretrained on ImageNetv2~\cite{recht2019imagenet}, CLIP~\cite{narasimhan2021clip} pretrained on DataComp-1B~\cite{gadre2023datacomp}, DINO~\cite{caron2021emerging} pretrained on Google Landmarks v2~\cite{weyand2020google}, and DINOv2 pretrained on LVD-142M~\cite{oquab2023dinov2} which is a composite dataset introduced by the same authors.

For each shot $s_i$, we construct a set of candidate segments from the broadcast video using a sliding window of length $l_i$ and stride 1, defined as:
\begin{equation}
b_j^{(i)} = \{\mathbf{g}_j, \mathbf{g}_{j+1}, \dots, \mathbf{g}_{j + l_i - 1}\}, \quad j = 1, \dots, N_{b} - l_i + 1.
\label{eq:sliding_window}
\end{equation}
Let \( \phi(s_i) \) and \( \phi(b_j^{(i)}) \) denote the temporally averaged feature representations of the summary shot \( s_i \) and each corresponding candidate broadcast segment \( b_j^{(i)} \), respectively. We measure the distance $d_{ij}$ between the averaged features as:

\begin{equation}
d_{ij} = \left\lVert\phi(s_i) - \phi(b_j^{(i)})\right\rVert_{2} ,
\label{eq:euclidean_distance}
\end{equation}
where $\left\lVert\cdot\right\rVert_2$ denotes the Euclidean norm. 
The best aligned segment from the broadcast video to the $i$-th summary shot, $b_{j^*}^{(i)}$, is then selected as:
\begin{equation}
b_{j^*}^{(i)} = \arg\min_j d_{ij}.
\label{eq:nearest_neighbor}
\end{equation}
Finally, correspondences are manually refined to fix any gaps or misalignments.

\begin{table*}[h]
\centering
\caption{Distribution of summary content (in \%) by match phase across different seasons in the Spanish, French, and Italian leagues. Additionally, the average per phase weighted by the number of games per league and season is shown.}
\label{tab:summary_by_season_and_league}
\resizebox{\textwidth}{!}{%
\begin{tabular}{@{}lccc|ccc|ccc|c@{}}
\toprule
& \multicolumn{3}{c|}{\textbf{Spain (La Liga)}} 
& \multicolumn{3}{c|}{\textbf{France (Ligue 1)}} 
& \multicolumn{3}{c|}{\textbf{Italy (Serie A)}} 
& \multirow{2}{*}{\textbf{Weighted Avg.}} \\
\cmidrule(lr){2-4} \cmidrule(lr){5-7} \cmidrule(lr){8-10}
\textbf{Match phase} & \textbf{2014/15} & \textbf{2015/16} & \textbf{2016/17} 
              & \textbf{2014/15} & \textbf{2015/16} & \textbf{2016/17} 
              & \textbf{2014/15} & \textbf{2015/16} & \textbf{2016/17} 
              &  \\
\midrule
Pre-Match      & 3.91  & 6.18  & 3.21  & 5.60  & 7.28  & 13.70  & 12.09  & 13.99  & 9.94  &  7.81 \\
First Half     & 41.53 & 39.78 & 37.73 & 46.47 & 34.59 & 31.40 & 41.50 & 41.50 & 38.97 &  38.00 \\
Half-Time      & 0.51  & 0.95  & 0.35  & 7.06  & 5.97  & 7.73 & 2.32  & 2.32 & 1.51  &  2.00 \\
Second Half    & 49.38 & 47.41 & 50.65 & 40.83 & 45.79 & 39.32 & 36.61 & 36.61 & 44.11 &  45.76 \\
Post-Match     & 4.67  & 5.69  & 8.06  & 0.00  & 6.38  & 7.85  & 5.59  & 5.59  & 5.47  &  6.42 \\
\bottomrule
\end{tabular}%
}
\end{table*}

\subsection{Description} \label{subsection:description_dataset}

After the pairing process, some considerations must be taken into account. Although the vast majority of plays are present in the broadcast videos, certain aspects of live production can affect their alignment with the summary. For example, in the case of replays, the camera angles used in the summary may differ from those in the broadcast footage. 

During the manual refinement stage, we address these discrepancies by aligning plays with their most accurate counterparts, even when captured from different viewpoints. Similarly, scenes such as crowd reactions or final greetings may not appear identically in both versions; in these cases, we pair them with the most visually and contextually similar segments based on human judgment.
Conversely, out-of-context shots with minimal relevance to the match storytelling, present in the summaries but absent from the live broadcast, are discarded to avoid confusion and maintain coherence in the annotated dataset.

% \begin{table*}[ht]
% \centering
% \caption{Distribution of games per split, categorized by league and season.}
% \label{tab:split_by_season_and_league}
% \resizebox{\textwidth}{!}{%
% \begin{tabular}{@{}lccc|ccc|ccc|c@{}}
% \toprule
% & \multicolumn{3}{c|}{\textbf{Spain (La Liga)}} 
% & \multicolumn{3}{c|}{\textbf{France (Ligue 1)}} 
% & \multicolumn{3}{c|}{\textbf{Italy (Serie A)}} 
% & \multirow{2}{*}{\textbf{Total}} \\
% \cmidrule(lr){2-4} \cmidrule(lr){5-7} \cmidrule(lr){8-10}
% \textbf{Split} & \textbf{2014/15} & \textbf{2015/16} & \textbf{2016/17} 
%               & \textbf{2014/15} & \textbf{2015/16} & \textbf{2016/17} 
%               & \textbf{2014/15} & \textbf{2015/16} & \textbf{2016/17} 
%               &  \\
% \midrule
% Train      & 6  & 18  & 42  & 1  & 2  & 23  & 5  & 4  & 40  &  141 \\
% Validation     & 6 & 9 & 11 & 0 & 1 & 8 & 3 & 1 & 12 &  51 \\
% Test      & 6  & 8  & 10  & 0  & 0  & 2 & 1  & 2 & 16  &  45 \\
% \midrule
% \textit{Total Games}     & 18    & 35    & 63    & 1     & 3     & 33    & 9     & 7     & 68    & 237 \\
% \bottomrule

% \end{tabular}%
% }
% \end{table*}

\begin{table}[htb]
\centering
\caption{Distribution of games in the proposed dataset, organized by split (training, validation, test), league, and season. For each league–season combination, the average duration of the summary videos (in minutes) is reported.}
\label{tab:split_by_league_and_season}
\resizebox{\columnwidth}{!}{%
\begin{tabular}{@{}llccc|c||c@{}}
\toprule
\textbf{League} & \textbf{Season} & \textbf{Train} & \textbf{Validation} & \textbf{Test} & \textbf{Total} & \textbf{Avg. Length} \\
\midrule
Spain  & 2014/15 & 6  & 6  & 6  & 18 & 5:46 \\
       & 2015/16 & 18 & 9  & 8  & 35 & 3:19 \\
       & 2016/17 & 42 & 11 & 10 & 63 & 1:41 \\
\midrule
France & 2014/15 & 1  & 0  & 0  & 1  & 3:37 \\
       & 2015/16 & 2  & 1  & 0  & 3  & 3:31 \\
       & 2016/17 & 23 & 8  & 2  & 33 & 3:38 \\
\midrule
Italy  & 2014/15 & 5  & 3  & 1  & 9  & 3:51 \\
       & 2015/16 & 4  & 1  & 2  & 7  & 3:51 \\
       & 2016/17 & 40 & 12 & 16 & 68 & 3:53 \\
\midrule\midrule
\textit{Total} &        & 141 & 51 & 45 & 237 & 3:19 \\
\bottomrule
\end{tabular}%
}
\end{table}

To divide the games into training, validation, and test sets, we adopt the split defined by SoccerNet, ensuring that each match in our dataset belongs to the same subset as in the original benchmark. The distribution of the games in these sets is shown in Table~\ref{tab:split_by_league_and_season}. We maintain a balanced organization, allocating $141$ games for training, $51$ for validation, and $45$ for testing. Each split includes multiple representations from different leagues and seasons, except for the French $2014/15$ and $2015/16$ seasons, which are underrepresented due to limited availability. From Table~\ref{tab:split_by_league_and_season}, we observe that the matches that are the most represented come from the $2016/17$ seasons of the Spanish and Italian leagues. This is beneficial as each league follows distinct editorial conventions, ensuring that each subset includes diverse summary styles and enhancing the generalization potential of models trained on these data. This is reflected in the average length of the summaries. Italy and France exhibit consistent durations throughout the seasons, approximately $3$ minutes and $30$ seconds for France, and $3$ minutes and $50$ seconds for Italy. In contrast, the Spanish league shows significant variation, with summaries exceeding $5$ minutes in the $2014/15$ season and dropping to under $2$ minutes by $2016/17$. Such variability highlights the challenges of the summarization task, as it is heavily influenced by editorial preferences, which can shift considerably over time and across leagues.

To analyze this variation, Table~\ref{tab:summary_by_season_and_league} presents the summary proportions broken down by game phase. We divide each match into two playable segments (first and second halves) and three non-playable phases (pre-match, half-time, and post-match). 
Table~\ref{tab:summary_by_season_and_league} highlights the differing editorial styles by illustrating the relative emphasis placed on each phase. On average, approximately $84\%$ of the content in the summaries corresponds to actual game-play, which means that around $16\%$ consists of clips unrelated to the plays themselves. The editorial differences between leagues are particularly evident in the half-time phase. Spanish summaries rarely include content from this interval, whereas Italian and French summaries dedicate roughly $2\%$ and $7\%$ of their duration to it, respectively. This can be extrapolated to the \textit{Pre-Match} phase, where significant differences are observed between leagues and seasons, whereas the \textit{Post-Match} phase is treated in a more consistent manner across all leagues.

\subsection{Experiments} \label{subsection:experiments_dataset}

To assess which of the proposed approaches most effectively optimizes the semi-automatic summary annotation process, we compare each produced output before manual refinement against the final ground truth alignments obtained through human supervised annotation. The comparison is conducted at the frame level by inferring frame indices from the segment boundary timestamps, using a sampling rate of $2$ frames per second (fps). Evaluation metrics include precision, recall, and F1 score per frame, where positive frames are those proposed after the alignment stage as defined by Equation~\eqref{eq:nearest_neighbor}. Additionally, we compute the Intersection over Union (IoU) between the set of frames proposed by the algorithm and those manually annotated, quantifying the proportion of correctly aligned frames relative to their union.

% \begin{table}[ht]
% \centering
% \caption{Performance comparison of KNN and TransNet across different backbones, showing Precision, Recall, F1 Score, and IoU (in \%).}
% \label{tab:knn_transnet_backbones_comparison}
% \resizebox{\columnwidth}{!}{%
% \begin{tabular}{@{}lcccc|cccc@{}}
% \toprule
% & \multicolumn{4}{c|}{\textbf{KNN-based}} 
% & \multicolumn{4}{c}{\textbf{TransNetv2}} \\
% \cmidrule(lr){2-5} \cmidrule(lr){6-9}
% \textbf{Metric} & \textbf{ResNet} & \textbf{CLIP} & \textbf{DINO} & \textbf{DINOv2} 
%                 & \textbf{ResNet} & \textbf{CLIP} & \textbf{DINO} & \textbf{DINOv2} \\
% \midrule
% Precision & 81.0 & 83.5 & 82.7 & 84.3 & 88.9 & 89.5 & 90.1 & 91.2 \\
% Recall    & 79.5 & 80.8 & 81.1 & 82.0 & 87.7 & 88.4 & 89.0 & 90.3 \\
% F1 Score  & 80.2 & 82.1 & 81.9 & 83.1 & 88.3 & 88.9 & 89.5 & 90.7 \\
% IoU       & 69.0 & 71.2 & 70.8 & 72.0 & 76.5 & 77.3 & 78.1 & 79.4 \\
% \bottomrule
% \end{tabular}%
% }
% \end{table}

\begin{table}[htb]
\centering
\caption{Evaluation of pair alignment performance using different shot segmentation strategies (SBD) and feature extraction backbones based on frame-level metrics.}
\label{tab:knn_transnet_backbones}
\resizebox{\columnwidth}{!}{%
\begin{tabular}{@{}ll||cccc@{}}
\toprule
\textbf{SBD} & \textbf{Backbone} & \textbf{Precision} & \textbf{Recall} & \textbf{F1 Score} & \textbf{IoU} \\
\midrule
KNN     & ResNet  & 0.2195 & 0.0725 & 0.1090 & 0.0577 \\
        & CLIP    & 0.7668 & 0.7310 & 0.7485 & 0.5980 \\
        & DINO    & 0.8169 & 0.7800 & 0.7980 & 0.6640 \\
        & DINOv2  & \textbf{0.8578} & \textbf{0.8244} & \textbf{0.8407} & \textbf{0.7252} \\
\midrule\midrule
TransNet & ResNet  & 0.0824 & 0.0337 & 0.0478 & 0.0245 \\
$th=0.5$      & CLIP    & 0.4463 & 0.3924 & 0.4176 & 0.2639 \\
         & DINO    & 0.5588 & 0.5046 & 0.5303 & 0.3609 \\
         & DINOv2  & 0.4458 & 0.3933 & 0.4179 & 0.2641 \\
\midrule
TransNet & ResNet  & 0.2326 & 0.0844 & 0.1238 & 0.0660 \\
$th=0.05$     & CLIP    & 0.7556 & 0.7039 & 0.7288 & 0.5733 \\
         & DINO    & 0.7518 & 0.7025 & 0.7263 & 0.5703 \\
         & DINOv2  & 0.7872 & 0.7373 & 0.7615 & 0.6148 \\
\bottomrule
\end{tabular}%
}
\end{table}

Although TransNetv2 performs well in detecting abrupt transitions for boundary segmentation, it struggles to generate reliable segment proposals for subsequent alignment, as shown in Table~\ref{tab:knn_transnet_backbones}. Applying oversegmentation with a $0.05$ threshold improves pairing performance, but it still underperforms compared to our custom kNN-based segmentation approach, which consistently outperforms TransNetv2 across all evaluated metrics.
Regarding the backbone networks used for frame-level feature extraction, the widely adopted ResNet-152 performs poorly relative to other candidates, producing suboptimal representations for shot comparison. Despite the fact that DINO is pretrained in a domain closer to our task, its successor, DINOv2, achieves significantly better results, reaching a notable F1 score of 0.8407 with our kNN-based approach. Consequently, DINOv2 enables a considerable speed-up in the annotation process compared to all other evaluated backbones.

To quantify the amount of effort required from an editor to annotate a new game, we propose a shot-level evaluation based on the best configuration reported in Table~\ref{tab:knn_transnet_backbones}. A predicted shot is labeled as positive if its IoU score with a ground truth shot exceeds the specified threshold in Table~\ref{tab:segment_metrics}. 
In this setting, recall measures the proportion of correctly identified summary shots. Emphasizing recall is crucial, as it is considerably more challenging for a human to detect missing segments in the broadcast video than to reject irrelevant proposals that do not belong in the summary.
Following Table~\ref{tab:segment_metrics} (a), both precision and recall tend to decrease as the IoU threshold increases. This is primarily due to the imperfect temporal alignment of the segments within the dataset. It is important to note that more than $70\%$ of the segments achieve at least a $0.45$ IoU, which serves as a strong indicator for subsequent manual verification. For a human annotator, having a rough indication of where a coincidence might occur is extremely valuable, as it eliminates the need to search throughout the duration of the match. Even a minimal overlap can significantly ease the alignment task.

Notably, at the $0.05$ IoU threshold, our method successfully captures $92.5\%$ of the relevant segments in the dataset. This shows that our approach can substantially reduce manual effort and annotation time by providing coarse but effective alignment cues.

Furthermore, in Table~\ref{tab:segment_metrics} (b), we introduce a time tolerance margin around the segment boundaries for pairs that do not meet the IoU overlap criterion. This adjustment acknowledges that proximity to the correct segment, although not overlapping, can still be highly informative. Given the inherent limitations in achieving perfect alignment, as discussed in Section~\ref{subsection:description_dataset}, it should be noted that when a 1-minute tolerance window is applied around each target segment, the recall increases from 0.9258 to 0.9857. This suggests that nearly all missing relevant segments can be recovered by inspecting just the surrounding minute of the match rather than reviewing the entire game, which offers substantial practical value to annotators.

\begin{table}[htb]
\centering
\caption{Precision and recall across IoU thresholds (a) and time tolerances (b) in a shot-level evaluation.}
\label{tab:segment_metrics}
\begin{subtable}[t]{0.48\columnwidth}
\centering
\caption{IoU thresholds}
\begin{tabular}{@{}ccc@{}}
\toprule
\textbf{IoU} & \textbf{Precision} & \textbf{Recall} \\
\midrule
0.05 & 0.7649 & 0.9256 \\
0.15 & 0.7341 & 0.9136 \\
0.25 & 0.7101 & 0.8970 \\
0.35 & 0.6733 & 0.8618 \\
0.45 & 0.5949 & 0.7174 \\
0.55 & 0.5450 & 0.6692 \\
0.65 & 0.5450 & 0.6132 \\
0.75 & 0.4805 & 0.5407 \\
0.85 & 0.3642 & 0.4098 \\
0.95 & 0.1567 & 0.1763 \\
\bottomrule
\end{tabular}
\end{subtable}
\hfill
\begin{subtable}[t]{0.48\columnwidth}
\centering
\caption{Time tolerances (in seconds)}
\begin{tabular}{@{}ccc@{}}
\toprule
\textbf{t (s)} & \textbf{Precision} & \textbf{Recall} \\
\midrule
0   & 0.7709 & 0.9258 \\
10  & 0.8180 & 0.9556 \\
20  & 0.8366 & 0.9682 \\
30  & 0.8474 & 0.9753 \\
40  & 0.8558 & 0.9809 \\
50  & 0.8619 & 0.9837 \\
60  & 0.8687 & 0.9857 \\
70  & 0.8729 & 0.9877 \\
80  & 0.8774 & 0.9890 \\
90  & 0.8817 & 0.9896 \\
\bottomrule
\end{tabular}
\end{subtable}
\end{table}

%% Model
\section{Baseline Model}\label{section:model}

To obtain summaries from soccer game videos, we propose a baseline architecture designed as a benchmark for future research on soccer video summarization. It is detailed in Section~\ref{subsection:pipeline_model} while Section~\ref{section:experiments_model} provides experimental results and ablations with this baseline model.

\subsection{Pipeline}\label{subsection:pipeline_model}

\begin{figure*}[htbp]
  \centering
  \includegraphics[width=\textwidth]{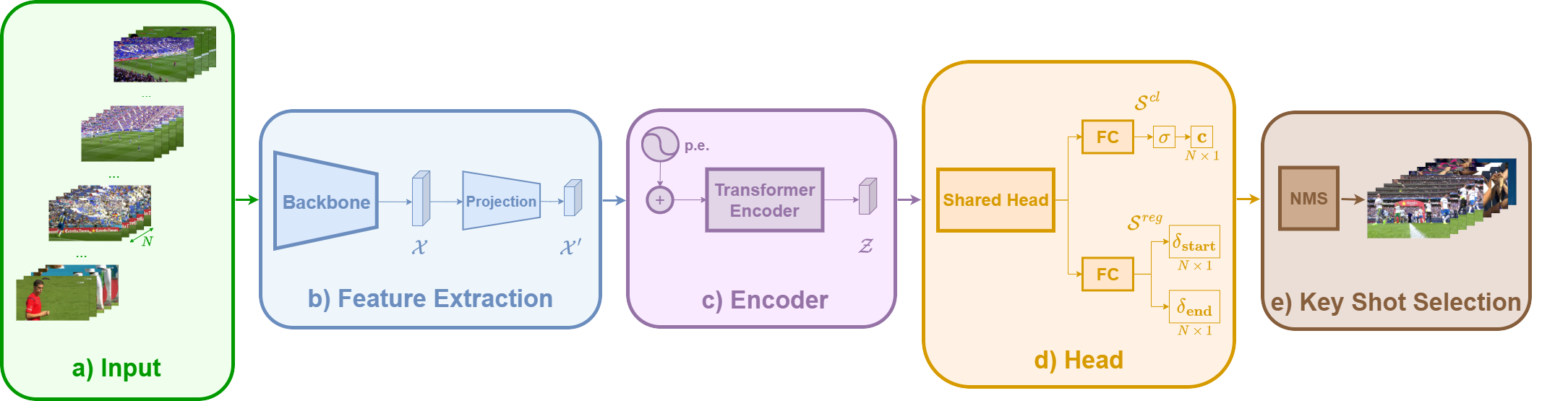}
  \Description{Diagram illustrating video summarization process.}
  \caption{Baseline model overview. Chunks of $N$ frames are extracted from the full-match broadcast (a) and are first passed through a feature extraction stage (b), where each frame is represented by a feature vector and projected into a shared embedding space. Sinusoidal positional encoding (p.e.) is then added before feeding the sequence into a Transformer encoder (c). The confidence score $c$ and the boundary offsets $\delta_{\text{start}}$ and $\delta_{\text{end}}$ are obtained through a shared prediction head (d). During inference (e), Non-Maximum Suppression (NMS) is applied to produce the final set of shot proposals included in the summary.}
  \label{fig:video_summarization}
\end{figure*}

As shown in Figure~\ref{fig:video_summarization}, the architecture consists of three main components: a feature extraction stage, a Transformer encoder, and a classification head.
The model processes input sequences segmented into fixed-length chunks extracted from the full broadcast video, following the approach adopted in previous soccer video understanding works such as \cite{cartas2024two, cartas2022graph}. This chunk-based approach reduces computational cost by avoiding the need to process the entire match at once and instead focuses on shorter segments, which are more manageable for identifying shots suitable for the final summary.

We define the number of frames per chunk as the length of the chunk in seconds multiplied by the frame rate as frames per second, that is,
\begin{equation}
\label{eq:chunk_size}
N = chunk~size~(s) \cdot frame~rate~(fps)
\end{equation}

\paragraph{\textbf{Feature Extraction}}
In this stage, we construct a feature vector $\mathbf{X} \in \mathbb{R}^{B \times N \times d_{\text{in}}}$, where $B$ is the batch size, $N$ is the number of frames per chunk, and $d_{\text{in}}$ denotes the dimensionality of the extracted features. These per-frame representations are obtained using backbone networks pretrained in large-scale visual recognition tasks, with the goal of capturing spatial and short-term visual cues while preserving the temporal structure of the input sequence.

To align all features in a common representation space, a linear projection is applied, resulting in a transformed tensor $\mathbf{X'} \in \mathbb{R}^{B \times N \times d}$, where $d$ is the embedding dimension used throughout the model. These projected features are then fed into the Transformer encoder, which is responsible for modeling long-range dependencies within each chunk and identifying the frames that are most relevant for summary inclusion.

\paragraph{\textbf{Encoder}}
Once the features are projected, a sinusoidal positional encoding is added to $\mathbf{X'}$ to encode temporal ordering. 
The resulting sequence of embeddings is passed through a two-layer Transformer encoder, initialized from scratch, to model long-range dependencies across the input sequence, due to the multi-head attention mechanism, obtaining an output tensor $\mathbf{Z} \in \mathbb{R}^{B \times N \times d}$.

\paragraph{\textbf{Head}}
The proposed prediction head adopts the anchor-free formulation introduced in \cite{dnset2021}, integrating both the classification and regression components, avoiding the centerness score. It is built on a set of shared layers comprising a series of linear projections with ReLU activations~\cite{nair2010rectified}, dropout, and layer normalization. The resulting shared representation is then passed to two parallel task-specific heads: one for per-frame classification and the other for per-frame boundary offset regression.

For classification, we adopt a per-frame supervision strategy based on annotated summary intervals $\mathbf{Y}^{cl} \in \{0, 1\}^{B \times N \times 1}$, enabling the model to identify the frames that contribute the most to the final summary. The classification head consists of a linear layer with sigmoid activation ($\sigma$) that gives a score confidence $\mathbf{S}^{cl} \in [0, 1]^{B \times N \times 1}$.

In parallel, a regression head is defined as a linear layer that maps the input representation to a 2-channel output prediction $\mathbf{S}^{\text{reg}} \in \mathbb{R}^{B \times N \times 2}$ to estimate the boundary offsets for summary shots. The ground truth offsets $\mathbf{Y}^{\text{reg}} \in \mathbb{R}^{B \times N \times 2}$ are derived from the discontinuities in the annotated intervals, representing the temporal distance from the current frame to the shot boundaries ($\delta_{\text{start}}, \delta_{\text{end}}$). Frames that do not belong to the summary are assigned an offset of $-1$ to indicate irrelevance.

\paragraph{\textbf{Key shot selection}} 
During inference, we apply Non-Maximum Suppression (NMS) to perform key shot selection based on the predicted classification scores and boundary offsets. This step filters out redundant overlapping proposals while preserving only the most confident predictions to ensure a coherent final summary. For each frame, the boundary positions are computed from the predicted offsets, $\delta_{\text{start}}$ and $\delta_{\text{end}}$, as 
\begin{equation}
\label{eq:boundaries}
n_{\text{start}} = n - \delta_{\text{start}}, \quad n_{\text{end}} = n + \delta_{\text{end}},
\end{equation}
where $n$ is the frame index. 
Among all overlapping shot proposals from frames with a classification score $c$ above a threshold of 0.5, the one with the highest score is selected, and the remaining candidates are discarded. The selection after the NMS stage, both the shot boundaries and the classification score, constitutes the shot selection proposal for the input chunk.

\paragraph{\textbf{Loss}}
The model is trained using a combination of loss terms, as defined in Equation~\eqref{eq:loss}, consisting of a weighted sum of a classification loss, $\mathcal{L}{\text{cls}}$, which determines whether each frame belongs to the summary, and a regression loss, $\mathcal{L}{\text{reg}}$, which estimates the boundary offsets of each positive frame relative to the start and end of its corresponding shot, as predicted by the regression head. More specifically, $\mathcal{L}_{\text{cls}}$ is defined as a binary cross-entropy:

\begin{equation}
\mathcal{L}_{\text{cls}} = - \frac{1}{B \cdot N} \sum_{b=1}^{B} \sum_{n=1}^{N} \left[
Y^{\text{cls}}_{b,n} \cdot \log S^{\text{cls}}_{b,n} + (1 - Y^{\text{cls}}_{b,n}) \cdot \log (1 - S^{\text{cls}}_{b,n})
\right].
\end{equation}
On the other hand,  $\mathcal{L}_{\text{reg}}$ is based on the following smooth $L_1$ loss: 
\begin{equation}
\ell_{b,n,c} =
\begin{cases}
0.5 \cdot (S^{\text{reg}}_{b,n,c} - Y^{\text{reg}}_{b,n,c})^2, & \text{if } |S^{\text{reg}}_{b,n,c} - Y^{\text{reg}}_{b,n,c}| < 1 \\
|S^{\text{reg}}_{b,n,c} - Y^{\text{reg}}_{b,n,c}| - 0.5, & \text{otherwise.}
\end{cases}
\end{equation}
Thus, we have that 
\begin{equation}
\mathcal{L}_{\text{reg}} = \frac{1}{M} \sum_{b=1}^{B} \sum_{n=1}^{N} \sum_{c=1}^{2} \mathds{1}_{Y^{\text{reg}}_{b,n,c} \geq 0} \cdot \ell_{b,n,c}
\end{equation}
where $\mathds{1}$ is the indicator function and $M$ denotes the number of positive frames, corresponding to those that fall within the ground truth summary segments, that is:
\begin{equation}
M = \sum_{b=1}^{B} \sum_{n=1}^{N} \sum_{c=1}^{2} \mathds{1}_{Y^{\text{reg}}_{b,n,c} \geq 0}
\end{equation}
Finally, the total loss is:
\begin{equation}\label{eq:loss}
\mathcal{L} = \mathcal{L}_{\text{cls}} + \lambda \cdot \mathcal{L}_{\text{reg}}
\end{equation}
where $\lambda\geq 0$ controls the contribution of the regression loss.

\paragraph{\textbf{Implementation Details}} For training, the feature extraction backbone is kept frozen, and the features are precomputed offline. Using these features, we train the rest of the model with the AdamW optimizer~\cite{loshchilov2017decoupled}. The training setup includes a learning rate of $\gamma = 5 \cdot 10^{-5}$, $\beta_1 = 0.9$, $\beta_2 = 0.999$, a weight decay of 0.05, and a batch size of 64 chunks. The network is trained using a learning rate plateau scheduler with patience of 6 epochs and a maximum of 150 epochs. To improve generalization and expose the model to more challenging examples, the MixUp~\cite{zhang2017mixup} augmentation is applied only to the training set, with parameters sampled from a beta distribution where $\alpha = 0.3$ and $\beta = 0.3$.

\subsection{Ablation experiments}\label{section:experiments_model}

We conduct our experiments using the ground truth derived from the manually refined annotations and organize the data according to the predefined splits, all described in Section~\ref{section:Dataset}. The validation subset is used during training for the optimal model selection, while the test subset is reserved exclusively for final evaluation. All results presented in this section are computed in the test set using per-frame evaluation metrics.

We present a series of experiments aimed at analyzing, through an ablation study, the impact of various design choices on the performance of the proposed baseline model for soccer video summarization.

\paragraph{\textbf{Feature extraction}} The first study explores how the choice of backbone for frame-level feature extraction impacts model performance. This stage is critical, as it determines the model’s ability to capture both semantic content and local temporal relationships. We evaluate two widely used 2D backbones, CLIP (pretrained on DataComp-1B) and ResNet-152 (pretrained on ImageNetv2), and compare them with a transformer-based video model, VideoMAEv2 (pretrained on Kinetics-710). While CLIP and ResNet-152 extract semantic features from individual frames, VideoMAEv2 processes short clips, allowing it to capture both semantic cues and short-term temporal dynamics. Specifically, each frame representation extracted from the VideoMAEv2 encoder is constructed using a sliding window that includes its immediate neighboring frames. We experiment with two versions of the VideoMAEv2 encoder: ViT-Small and ViT-Giant.

\begin{table}[htb]
\centering
\caption{Comparison of different backbone architectures. Results are reported in terms of evaluation metrics and the number of parameters for each backbone.}
\label{tab:ablation_backbone}
\resizebox{\columnwidth}{!}{%
\begin{tabular}{@{}lccc||c@{}}
\toprule
\textbf{Backbone} & \textbf{Precision} & \textbf{Recall} & \textbf{F1 Score} & \textbf{\# params} \\
\midrule
ResNet-152     & 0.3603 &  0.2145 & 0.2689 & 58.1M \\
CLIP     &  0.5048 & 0.2335 & 0.3193 & 151.3M \\
VideoMAEv2 - ViT small    & 0.4426 & 0.2797 & 0.3428 & 21.9M \\
VideoMAEv2 - ViT giant & 0.4796 & 0.3367 & \textbf{0.3956} & 1011.6M \\
\bottomrule
\end{tabular}%
}
\end{table}

We evaluate the performance in the test set by computing the F1 score with a fixed chunk length of 60 seconds. As detailed in Table~\ref{tab:ablation_backbone}, the VideoMAEv2 giant encoder achieves the highest F1 score, outperforming all other configurations. This improvement is attributed to its capacity to model both spatial and short-term temporal dependencies effectively. Its larger number of parameters, compared to the smaller variant, further enhances generalization and enables the extraction of richer feature representations. Additionally, features extracted with CLIP surpass those obtained from ResNet‑152, suggesting that transformer-based architectures generalize more effectively to soccer video understanding tasks, in part due to their higher representational capacity.
The strong performance of VideoMAEv2 highlights the importance of explicitly modeling short-term temporal context, something that purely spatial backbones inherently lack.
Finally, the \textit{small} version of VideoMAEv2 offers a compelling trade-off between computational cost and performance, making it a suitable candidate for finetuning more domain-specific data, where efficiency and adaptability are critical.

% \begin{table}[ht]
% \centering
% \caption{Ablation study on chunk size effect. Performance is evaluated in terms of Precision, Recall, and F1 Score in the test set.}
% \label{tab:ablation_chunk_size}
% \resizebox{0.75\columnwidth}{!}{%
% \begin{tabular}{@{}cccc@{}}
% \toprule
% \textbf{Chunk Size (s)} & \textbf{Precision} & \textbf{Recall} & \textbf{F1 Score} \\
% \midrule
% 10  & 0.5193 & 0.1917 & 0.2800 \\
% 20  & 0.5545 & 0.2144 & 0.3093 \\
% 30  & 0.4192 & 0.2614 & 0.3220 \\
% 45  & 0.4897 & 0.2516 & 0.3324 \\
% 60  & 0.4426 & \textbf{0.2797} & \textbf{0.3428} \\
% 90  & 0.5404 & 0.2197 & 0.3124 \\
% 120 & 0.4828 & 0.2218 & 0.3039 \\
% 180 & \textbf{0.5553} & 0.1934 & 0.2869 \\
% \bottomrule
% \end{tabular}%
% }
% \end{table}

\begin{table}[htb]
\centering
\caption{Ablation study on the effect of chunk size.}
\label{tab:ablation_chunk_size}
\resizebox{0.75\columnwidth}{!}{%
\begin{tabular}{@{}cccc@{}}
\toprule
\textbf{Chunk Size (s)} & \textbf{Precision} & \textbf{Recall} & \textbf{F1 Score} \\
\midrule
15  & 0.4619 & 0.2946 & 0.3598 \\
30  & 0.4670 & 0.3079 & 0.3711 \\
45  & 0.4081 & 0.3742 & 0.3905 \\
60  & 0.4796 & 0.3367 & \textbf{0.3956} \\
75  & 0.4471 & 0.3514 & 0.3935 \\
90  & 0.4802 & 0.3296 & 0.3909 \\
105 & 0.4297 & 0.3379 & 0.3783 \\
120 & 0.4650 & 0.3114 & 0.3730 \\
\bottomrule
\end{tabular}%
}
\end{table}

\paragraph{\textbf{Chunk length}} Next, in Table~\ref{tab:ablation_chunk_size}, we analyze the impact of the duration of the chunk on the performance of the final key shot selection. Ranging from $15$ seconds chunks which can only focus in a very local context in terms of the game, to $2$ minutes chunks, which can keep a lot of contextual information around key actions. Optimal performance is achieved with $60$ second chunks, as shown in Table~\ref{tab:ablation_chunk_size}, where a turning point is observed in the F1 score. This trend suggests that increasing the duration of the chunk from $10$ to $60$ seconds progressively provides the model with more contextual information, improving its ability to identify relevant segments for summary. Between $45$ and $90$ seconds, the network maintains stable performance, as many key segments of the summary tend to fall within this temporal range. However, beyond this point, performance declines. This drop can be attributed to the nature of soccer dynamics. Longer chunks may introduce redundant content, such as extended general-view shots of the pitch, which can hinder the model's ability to focus on salient actions and make accurate decisions for summary selection.

\paragraph{\textbf{Heads}} Furthermore, we examine the contribution of each head to the final summary construction and analyze how the NMS strategy impacts performance when combining both heads. As described in the model section, our baseline employs a two-head design, integrating per-frame classification with boundary offset regression. Table~\ref{tab:ablation_nms} presents the F1 scores for three scenarios: using only the classification head with raw predictions to extract key frames; adding the regression head during training, but still relying on raw predictions for key frames; and finally, combining both heads with NMS applied during inference to refine the shot proposals.

\begin{table}[htb]
\centering
\caption{Comparison between classification-only and classification+regression heads, with or without NMS.}
\label{tab:ablation_nms}
\resizebox{\columnwidth}{!}{%
\begin{tabular}{@{}lcccc@{}}
\toprule
\textbf{Head} & \textbf{NMS} & \textbf{Precision} & \textbf{Recall} & \textbf{F1 Score} \\
\midrule
Classification & \ding{55} & 0.6450 & 0.2395 & 0.3493 \\
Classification + Regression & \ding{55} & 0.6041 & 0.2679 & 0.3712 \\
Classification + Regression & \ding{51} & 0.4796 & 0.3367 & \textbf{0.3956} \\
\bottomrule
\end{tabular}%
}
\end{table}

The results show that the performance of the model improves when both heads are combined with NMS, as indicated by the overall F1 score. Recall scores also follow this trend, suggesting improved generalization capabilities. In contrast, higher precision values are obtained when using only the classification head, which likely benefits from focusing on a limited set of concrete shot types (e.g., goals) rather than generating a broader summary representation.

\paragraph{\textbf{MixUp}} Finally, Table~\ref{tab:ablation_mixup} presents the impact of incorporating the MixUp strategy during training. As shown, this augmentation technique enhances generalization by mixing pairs of examples, leading to a higher F1 score.

\begin{table}[htb]
\centering
\caption{Ablation study evaluating the effect of MixUp data augmentation during training.}
\label{tab:ablation_mixup}
\resizebox{0.65\columnwidth}{!}{%
\begin{tabular}{@{}lccc@{}}
\toprule
\textbf{MixUp} & \textbf{Precision} & \textbf{Recall} & \textbf{F1 Score} \\
\midrule
\ding{55} & 0.4421 & 0.3115 & 0.3655 \\
\ding{51} & 0.4796 & 0.3367 & \textbf{0.3956} \\
\bottomrule
\end{tabular}%
}
\end{table}

To summarize, the best performance for the proposed baseline model is achieved under the following configuration: frame features are extracted using the giant encoder from VideoMAEv2; input sequences are divided into chunks of $60$~seconds; both the classification and regression heads are used; key shot selection during inference is carried out through Non-Maximum Suppression (NMS); and MixUp data augmentation is applied during training.

%% Evaluation
\section{Proposed metric and evaluation}

Although many video summarization methods \cite{dnset2021, apostolidis2020ac, he2023align}, constrain the generated summary to a fixed percentage of the total video duration, typically around $15\%$, we deliberately avoid imposing such a limitation. This decision allows the model greater flexibility to learn what constitutes a meaningful and representative summary. In the context of soccer, the content and duration of a summary can vary significantly depending on editorial choices, making rigid constraints less suitable.
Instead of restricting the model output, we constrain the evaluation by aligning the predicted summary length with that of the ground truth. During inference, predicted shots are ranked according to their importance scores, and only the top-ranked shots are selected until the cumulative duration matches the ground truth summary. This approach enables an objective evaluation by focusing on the model’s ability to identify key events, such as goals or pivotal moments, while ignoring more subjective content that can or cannot be included according to the editor's intent.
In this way, the model is encouraged to learn what makes a shot representative of the game without being conditioned to produce an output of a fixed length. This promotes better generalization and avoids penalizing the model for selecting relevant content that falls outside arbitrary constraints.

\begin{table}[htb]
\centering
\caption{Evaluation of key shot selection with metrics computed per game and macro-averaged across the test set. All metrics are calculated under the constraint that the predicted summary length matches that of the ground truth.}
\label{tab:evaluation}
\resizebox{0.75\columnwidth}{!}{%
\begin{tabular}{@{}lccc@{}}
\toprule
\textbf{Model} & \textbf{Precision$@$T} & \textbf{Recall$@$T} & \textbf{F1 Score$@$T} \\
\midrule
Baseline & 0.5034 & 0.3375 & 0.3883 \\
\bottomrule
\end{tabular}%
}
\end{table}

Building on this idea, we propose a new metric for objective evaluation in soccer video summarization. 
Rather than focusing exclusively on the presence of specific shots in the ground truth, we treat the ground truth summary as a representation of the most important moments of the match. We denote the temporal length of each ground truth summary as $T$ and define Precision$@T$, Recall$@T$, and the F1 Score$@T$ as evaluation metrics constrained by this length. As shown in Table~\ref{tab:evaluation}, the model achieves an F1 Score$@T$ of $0.3883$. This evaluation strategy allows for a more objective assessment of the generated content and, in combination with the proposed dataset, establishes a benchmark for future research in this area.

%% Conclusions
\section{Conclusions}\label{section:conclusions}

In this paper, we introduce a new curated public dataset for soccer video summarization, consisting of 237 pairs of summary and broadcast videos sourced from publicly available matches in SoccerNet. To support both the construction and future expansion of the dataset, we propose a semi-automated annotation pipeline that enables scalable alignment generation as more content becomes accessible.

Alongside the dataset, we present a baseline model specifically designed for the video summarization task. Through a comprehensive set of ablation studies, we evaluate the impact of various architectural choices, input configurations, and training strategies to identify the best-performing setup.

Together, the proposed dataset, baseline model, and tailored evaluation metrics establish the first dedicated benchmark for soccer video summarization, addressing a critical gap in the literature and providing a robust foundation for future research in this area.

%% Acknowledgements
\begin{acks}
This work was funded by the European Union (GA 101119800 - EMERALD).
The authors also acknowledge the EuroHPC Joint Undertaking for awarding us access to Karolina at IT4Innovations, Czech Republic and to MareNostrum5 at the Barcelona Supercomputing Center (BSC), Spain.

A. D.-J. would like to sincerely thank Dr. Alejandro Cartas for his invaluable support throughout this research.
\end{acks}

\bibliographystyle{ACM-Reference-Format}
\bibliography{references}

\end{document}